\documentclass[10pt,letterpaper]{article}

\usepackage{cogsci}
\usepackage{pslatex}
\usepackage[natbibapa]{apacite}
\usepackage{amsmath}
\usepackage{bm}
\usepackage{graphicx}

\title{Evidence for the size principle in semantic and perceptual domains}
 
\author{
{\large \bf Joshua C. Peterson (peterson.c.joshua@gmail.com)}\\
{\large \bf Thomas L. Griffiths (tom\_griffiths@berkeley.edu)} \\
Department of Psychology, University of California, Berkeley}

\begin{document}
\maketitle

\begin{abstract}
Shepard's Universal Law of Generalization offered a compelling case for the first physics-like law in cognitive science that should hold for all intelligent agents in the universe. Shepard's account is based on a rational Bayesian model of generalization, providing an answer to the question of \textit{why} such a law should emerge. Extending this account to explain how humans use multiple examples to make better generalizations requires an additional assumption, called the \textit{size principle}: hypotheses that pick out fewer objects should make a larger contribution to generalization. The degree to which this principle warrants similarly law-like status is far from conclusive. Typically, evaluating this principle has not been straightforward, requiring additional assumptions. We present a new method for evaluating the size principle that is more direct, and apply this method to a diverse array of datasets. Our results provide support for the broad applicability of the size principle.

\end{abstract}

\section{Introduction}

In the seminal work of \cite{shepard1987toward}, the notion of stimulus \textit{similarity} was made concrete through its interpretation as stimulus \textit{generalization}. It was shown that, across species (including humans), generalization probabilities follow an exponential law with respect to an internal psychological space. Specifically, the probability that $y$ is in some set $C$ that contains $x$ (what Shepard terms a ``consequential subset'') is an exponentially decreasing function of distance ($d$) in psychological space:
\begin{equation}
    s_{xy} = e^{-d(x,y)}.
\end{equation}
Shepard termed this phenomenon the Universal Law of Generalization, in that it should apply to any intelligent agent, anywhere in the universe. This result has been used in numerous cognitive models that invoke similarity (e.g., \citeauthor{nosofsky1986attention}, \citeyear{nosofsky1986attention}; \citeauthor{kruschke1992alcove}, \citeyear{kruschke1992alcove}).

In spite of this, one could argue that generalization from a single stimulus to another  does not adequately describe the full scope of human behavior. Indeed, in a concept learning task, people are asked to generalize from multiple examples of a concept. To capture this, \cite{tenenbaum2001generalization} extended Shepard's original Bayesian derivation of the law to rationally integrate information about multiple instances. The resulting model defines the probability of generalization (that $y$ is in $C$) as a sum of the probabilities of all hypotheses $h$ about the true set $C$ that include both $x$ and $y$,
\begin{equation}
    p(y \in C\mid x) = \sum_{h:y\in h} p(h \mid x).
\end{equation}
The posterior probability of each hypothesis is given by Bayes' rule,
\begin{equation}
    p(h \mid x) = \frac{p(x \mid h)p(h)}{p(x)}.
\end{equation}
The prior $p(h)$ represents the learner's knowledge about the consequential region before observing $x$. The likelihood $p(x \mid h)$ depends on our assumptions about how the process that generated $x$ relates to the set $h$. The key innovation over Shepard's model is the use of the likelihood function
\begin{equation}
    p(x \mid h) = \left \{ 
    \begin{array}{cc} 
    \frac{1}{|h|} & x \in h \\
    0 & \mbox{otherwise} 
    \end{array}
    \right .
\end{equation}
where $|h|$ is the number of objects in the set picked out by $h$.
The motivation for this choice of likelihood function is the \textit{size principle}, which uses the assumption of random sampling to justify the idea that smaller hypotheses should be given greater weight (see \citeauthor{tenenbaum2001generalization}, \citeyear{tenenbaum2001generalization}, for a demonstration of this when $x$ represents multiple examples). 

The value of the size principle lies in the fact that it allows for the benefit of multiple examples of a concept to influence generalization. Assuming samples are drawn independently, the likelihood of a hypothesis for $n$ samples is simply the likelihood of that hypothesis for a single sample to the power of $n$. From this, it can be shown that generalization tightens as the number of examples increases, consistent with human judgments \cite[see][]{tenenbaum2001generalization}.

\begin{figure*}[!ht]
\begin{center}
\includegraphics[trim={0 0 0 0},clip,width=\linewidth,keepaspectratio]{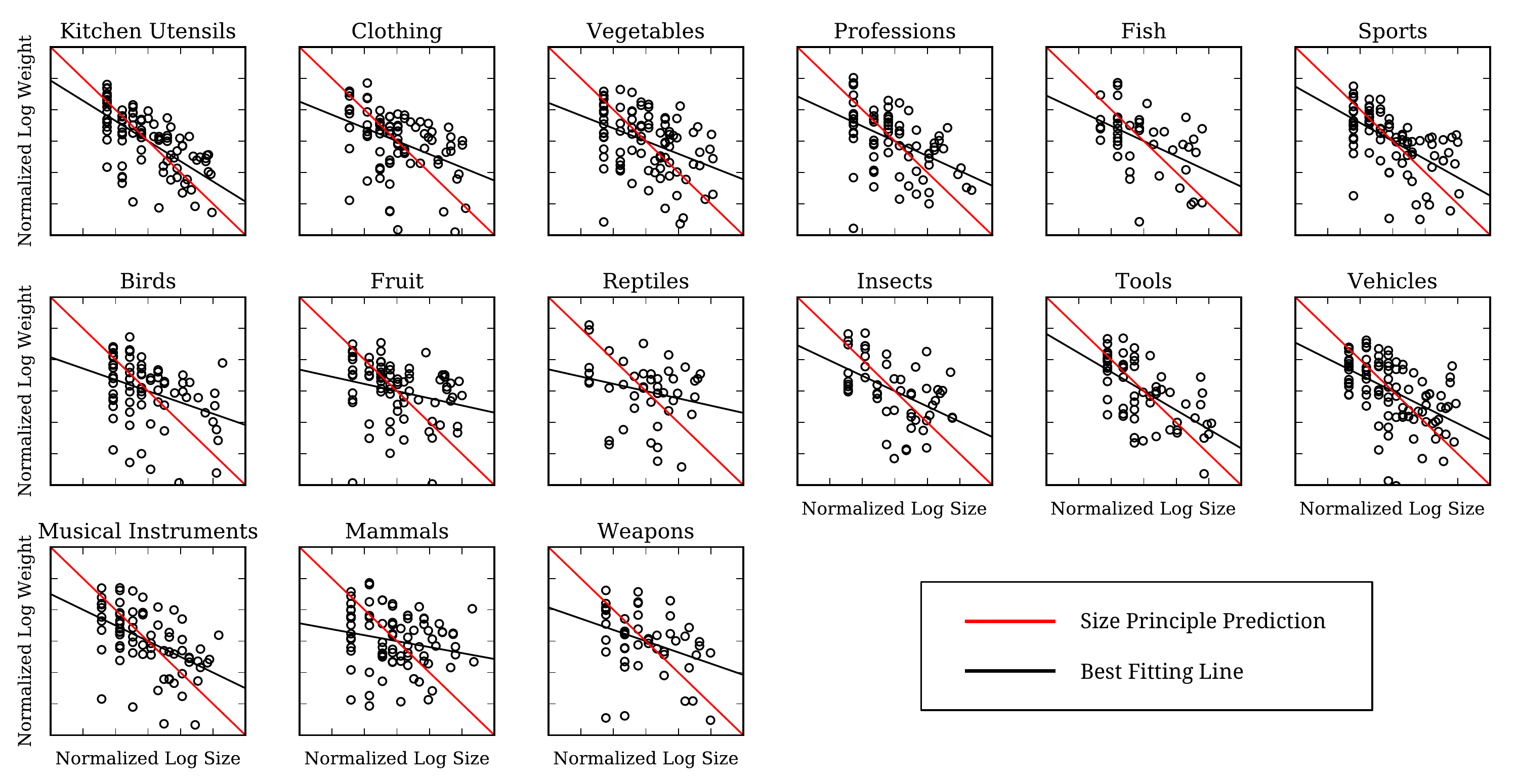}
\end{center}
\caption{Feature Size/Weight Relationships in Semantic Dataset Group 1.} 
\label{group1-plots}
\end{figure*}

The size principle thus plays an important role in understanding generalization, placing equal importance on determining whether it actually describes human similarity judgments in a wide range of settings. If the size principle is disconfirmed, an alternative augmentation of Shepard's model is needed to explain generalization from multiple instances. If it can be confirmed to hold broadly, it is a good candidate for a second law of generalization, or an amendment of the original law. In this paper we build on previous work evaluating the evidence for the size principle \citep{navarro2010similarity}, providing a novel and more direct methodology and a broader empirical evaluation that includes rich perceptual feature spaces. 

\section{Evaluating the Size Principle}

In this section we describe previous work evaluating the size principle and provide the details of our approach.

\subsection{Previous work}

\cite{navarro2010similarity} made three important contributions towards understanding the scope of the size principle. First, they made explicit a link between the Bayesian model of generalization and a classic model of similarity judgment. The similarities between a set of objects can be summarized in a similarity matrix ${\bf S}$, where the entry in the $i$th row and $j$th column gives the similarity $s_{ij}$ between objects $i$ and $j$. The additive clustering model \citep*{shepard1979additive} decomposes such a similarity matrix into the matrix product of a feature-by-object matrix ${\bf F}$, its transpose, and a diagonal weight matrix ${\bf W}$,
\begin{equation}
    {\bf S} = {\bf F}{\bf W} {\bf F}^T . \label{eq:adclus}
\end{equation}
The feature matrix $\bf F$ is binary and can represent any of a broad set of structures including partitions, hierarchies, and overlapping clusters, and can either be inferred by a number of different models or generated directly by participants. The individual entries of $\bf S$ are defined as
\begin{equation}
  \label{eq:entry-adclus}
  \begin{gathered}
\displaystyle s_{ij} = \sum_{k=1}^{N_{f}} w_{k}f_{ik}f_{jk} .
 \end{gathered}
 \end{equation}
 Navarro and Perfors (2010) pointed out that each feature could be taken as a single hypothesis $h$, as it likewise picks out a set of objects with a common property. Having made this link, the degree of generalization between objects $i$ and $j$ predicted by the Bayesian model can be put in the same format as Equation \ref{eq:entry-adclus}: a weighted sum of common features (a similar point was made by Tenenbaum \& Griffiths, 2001). The equivalence can be seen if we let $w_{k}$ represent the posterior $p(h \mid x)$ and $f_{k}$ be the $k^{\text{th}}$ hypothesis ($h_{k}$), since $f_{ik}f_{jk}$ selects only the features that contain both objects. If the prior probabilities of the different hypotheses are similar, the likelihood (and the size principle) will dominate and
\begin{equation}
  \label{eq:size_prin_w}
  \begin{gathered}
\displaystyle w_{k} \propto \frac{1}{|h_{k}|} .
 \end{gathered}
 \end{equation}
 
Using the link between hypotheses and features, Navarro and Perfors (2010) made their second contribution: an alternative derivation showing that the relationship predicted by the size principle can hold even in the absence of random sampling. They argued that learners encode the similarity structure of the world by learning a set of features $\bf F$ that efficiently approximate that structure.  Under this view, a ``coherent'' feature is said to be one for which all objects that possess that feature exhibit high similarity. If a learner seeks a set of features that are high in coherence, the size principle emerges even in the absence of sampling since the variability in the distribution of similarities between objects sharing a feature is a function of $|h_{k}|$. 

The third contribution that Navarro and Perfors (2010) made was to 
evaluate this prediction using data from the Leuven Natural Concept Database \cite[LNCD;][]{de2008exemplar}. This database consists of human-generated feature matrices for a large number of objects, as well as pair-wise similarity ratings for those objects. Navarro and Perfors (2010) observed that, under some simplifying assumptions, the size principle predicts that the similarity between objects that share a feature will be inversely related to the size of that feature. They showed that this prediction was borne out in 11 different domains analyzed in the LNCD. 

\begin{figure*}[!t]
\begin{center}
\includegraphics[trim={0 0 0 0},clip,width=\linewidth,keepaspectratio]{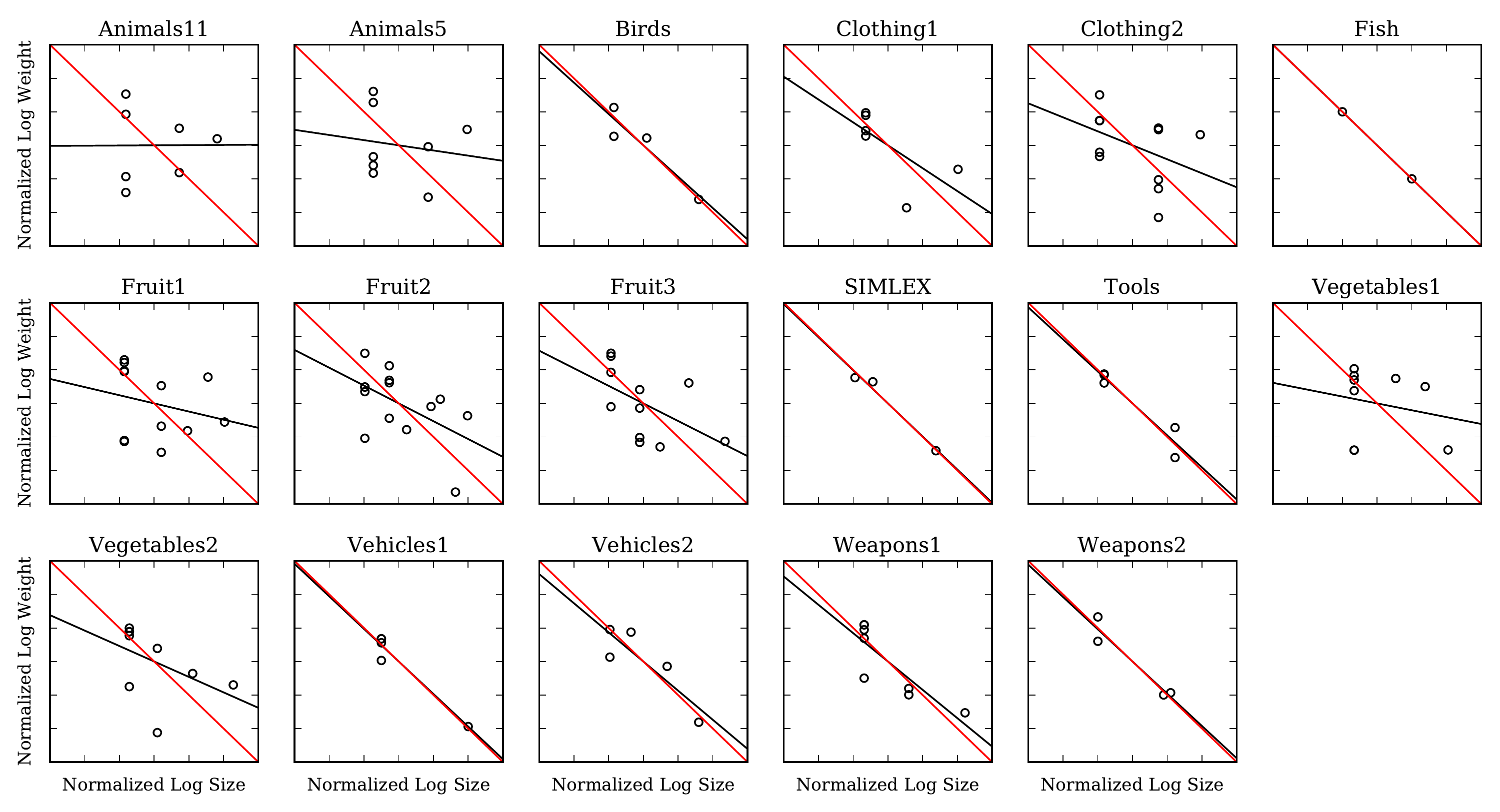}
\end{center}
\caption{Feature Size/Weight Relationships in Semantic Dataset Group 2.} 
\label{group2-plots}
\end{figure*}

\subsection{Directly testing the size principle}

The method adopted by Navarro and Perfors (2010) depends on the derived relationship between the similarity of objects that share a feature and the number of those objects. However, the link that they established between Bayesian clustering and the additive clustering model (Equations \ref{eq:adclus}-\ref{eq:size_prin_w}) can also be used to directly test the size principle. Since both models take the same mathematical form, we can directly test the size principle by estimating the weights $w_k$ for a set of features $\bf F$ and verifying that Equation \ref{eq:size_prin_w} holds. If we take the logarithm of both sides of this equation, we obtain the linear relationship
\begin{equation}
    \log w_k = -\log |h_{k}| + c
\end{equation}
which can be evaluated by correlating $w_k$ with the number of objects that possess feature $k$. Given a feature matrix ${\bf F}$ and similarity matrix ${\bf S}$, the weights $w_k$ can be obtained through linear regression \citep{peterson2016adapting}. In additive clustering the weights are often constrained to be non-negative. To obtain such weights, we employ a non-negative least squares algorithm \citep{lawson1995solving}. We can thus directly test the size principle in any domain where a feature matrix and a corresponding similarity matrix are available. In the remainder of the paper we consider two different sources of such matrices: semantic feature norms and perceptual neural networks.

\section{Semantic Hypothesis Spaces}

We first evaluate evidence for the size principle using two groups of datasets in which people judge the similarity of words. Both datasets contain human similarity ratings of noun pairs corresponding to concrete objects (e.g., ``zebra" and ``lion") and lists of binary feature labels associated with each object that can be filtered by frequency of mention. We call these features ``semantic" because they are linguistic descriptions of general concepts that exclude perceptual-level information associated with actual instances of that concept or brought to mind when the instance is perceived.
\subsection{Semantic Dataset Group 1} Following Navarro and Perfors (2010), the first evaluation dataset is comprised of similarity and feature matrices from the Leuven Natural Concept Database \citep{de2008exemplar}. It includes data for 15 categories (Kitchen Utensils, Clothing, Vegetables, Professions, Fish, Sports, Birds, Fruit, Reptiles, Insects, Tools, Vehicles, Musical Instruments, Mammals, and Weapons), each containing $\sim\!\!20-30$ exemplars. Binary feature matrices for each category contain $\sim\!\!200-300$ unique features each. The feature descriptions are much broader than merely visually apparent features (e.g., ``has wings", ``eats fruit", ``is attracted by shiny objects"). 
\subsection{Semantic Dataset Group 2} The second dataset group consisted of 17 similarity matrices from a variety of sources throughout the literature. The experimental contexts and methodologies differed considerably compared to the those in group 1. All but one of these datasets (SIMLEX) were taken from the similarity data repository on the website of \frenchspacing{Dr. Michael} Lee (http://faculty.sites.uci.edu/mdlee/similarity-data/). SIMLEX was taken from a larger word similarity dataset \citep{hill2016simlex}. The majority of the datasets (Birds, Clothing1, Clothing2, Fish, Fruit1, Fruit2, Furniture1, Furniture2, Tools, Vegetables1, Vegetables2, Weapons1, and Weapons2) are from \cite{romney1993predicting}. For dataset pairs such as (Vegetables1, Vegetables2), the first contains more prototypical items than the second. Since none of these datasets contain corresponding object-feature data, we matched objects from each set to the feature norms reported in \cite{mcrae2005semantic}.
\subsection{Analysis \& Results} For each dataset, we computed the element-wise multiplication of each pair of rows in $\bf F$ and used non-negative least squares to regress this matrix onto the corresponding empirical similarity values. We then computed the log of all non-zero weights, as well as the log of the feature sizes (column sums of the $\bf F$ matrix for which there was a corresponding non-zero weight). The resulting log weights and log feature sizes are z-score normalized and plotted in Figures \ref{group1-plots} and \ref{group2-plots} for each category in each subgroup. Red lines indicate perfect -1 slopes as predicted by the size principle, whereas black lines are best fitting lines to the actual data. The corresponding correlation coefficients are reported in Tables \ref{table-semantic1} and \ref{table-semantic2} along with a number of other statistics to be discussed.

Average Pearson and Spearman correlations were -0.43 and -0.47, respectively for group 1, and -0.63 and -0.61 for group2. For nearly all individual datasets in all groups, coefficients are consistently negative, with the exception of Animals11 in group 2, which along with Animals5 are the only datasets with no published method. Correlations were generally stronger for group 2. All correlations in group 1 were significant at the $\alpha = 0.05$ level except for the Reptiles and Mammals datasets. In contrast, virtually no correlations were significant in group 2 given the small number of features with non-zero coefficients, however one-sample \textit{t}-tests confirmed that the mean slopes were significantly less than 0 for both Pearson $(t(16)=-7.65, p<0.0001)$ and Spearman $(t(16)=-7.65, p<0.0001)$ correlations.

The $FR$ (feature ratio) column indicates how many coefficients were positive out of the total possible. Although there were many more features overall in group 1, the average percentage of features with non-zero weights was comparable (28\% and 23\% respectively). 

Finally, model performance in predicting similarity ($R^2$) is reported in the $R_{MP}^{2}$ column, and indicates the degree to which the feature sets are sufficient to accurately predict human similarity judgments. ($R^2$) values for group 1 are markedly higher than group 2 (which have many fewer features) and match reliability ceilings reported in the original experiments.

\begin{table}[!b]
\begin{center}
\caption{Correlations between feature size and feature weight (Semantic Datasets Group 1)}
\begin{tabular}{ l r r c r } 
\hline
\textbf{Set} & \textbf{Pearson} & \textbf{Spearman} & $\bm{FR}$ & $\bm{R^{2}_{MP}}$ \\ 
\hline
K. Utensils & -0.64 & -0.67 & 94/328 & 0.84 \\
Clothing & -0.42 & -0.47 & 84/258 & 0.71 \\
Vegetables & -0.41 & -0.43 & 91/291 & 0.68 \\
Professions & -0.48 & -0.51 & 73/370 & 0.76 \\
Fish & -0.48 & -0.49 & 43/156 & 0.80 \\
Sports & -0.58 & -0.65 & 85/382 & 0.81 \\
Birds & -0.36 & -0.37 & 72/225 & 0.75 \\
Fruit & -0.23 & -0.37 & 78/233 & 0.74 \\
Reptiles & -0.23 & -0.20 & 45/179 & 0.94 \\
Insects & -0.49 & -0.52 & 52/214 & 0.73 \\
Tools & -0.61 & -0.61 & 62/285 & 0.74 \\
Vehicles & -0.52 & -0.57 & 97/322 & 0.93 \\
M. Instruments & -0.50 & -0.56 & 72/218 & 0.90 \\
Mammals & -0.19 & -0.22 & 84/288 & 0.85 \\
Weapons & -0.36 & -0.38 & 49/181 & 0.88 \\
\hline
\end{tabular}
\label{table-semantic1}
\end{center}
\end{table}

\begin{table}[!b]
\begin{center}
\caption{Correlations between feature size and feature weight (Semantic Datasets Group 2)}
\vspace{2mm}
\begin{tabular}{ l r r c r } 
\hline
\textbf{Set} & \textbf{Pearson} & \textbf{Spearman} & $\bm{FR}$ & $\bm{R^{2}_{MP}}$ \\ 
\hline
Animals11 & 0.01 & 0 & 7/37 & 0.31 \\ 
Animals5 & -0.15 & -0.08 & 8/37 & 0.35 \\
Birds & -0.94 & -0.95 & 4/24 & 0.10 \\ 
Clothing1 & -0.68 & -0.78 & 6/28 & 0.10 \\ 
Clothing2 & -0.42 & -0.53 & 12/35 & 0.11 \\ 
Fish & -1.00 & -1.00 & 2/17 & 0.18 \\ 
Fruit1 & -0.24 & -0.29 & 12/38 & 0.19 \\ 
Fruit2 & -0.53 & -0.43 & 4/42 & 0.14 \\ 
Fruit3 & -0.52 & -0.64 & 11/42 & 0.25 \\ 
SIMLEX & -0.99 & -1.00 & 3/151 & 0.24 \\ 
Tools & -0.95 & -0.87 & 5/13 & 0.18 \\ 
Vegetables1 & -0.20 & -0.08 & 9/31 & 0.31 \\ 
Vegetables2 & -0.46 & -0.57 & 9/31 & 0.26 \\ 
Vehicles1 & -0.97 & -0.71 & 5/24 & 0.06 \\ 
Vehicles2 & -0.87 & -0.82 & 5/23 & 0.03 \\ 
Weapons1 & -0.85 & -0.87 & 8/32 & 0.16 \\ 
Weapons2 & -0.96 & -0.74 & 4/30 & 0.03 \\ 
\hline
\end{tabular}
\label{table-semantic2}
\end{center}
\end{table}

\begin{figure*}[!ht]
\begin{center}
\includegraphics[trim={37mm 0mm 81mm 0mm},clip,width=\linewidth,keepaspectratio]{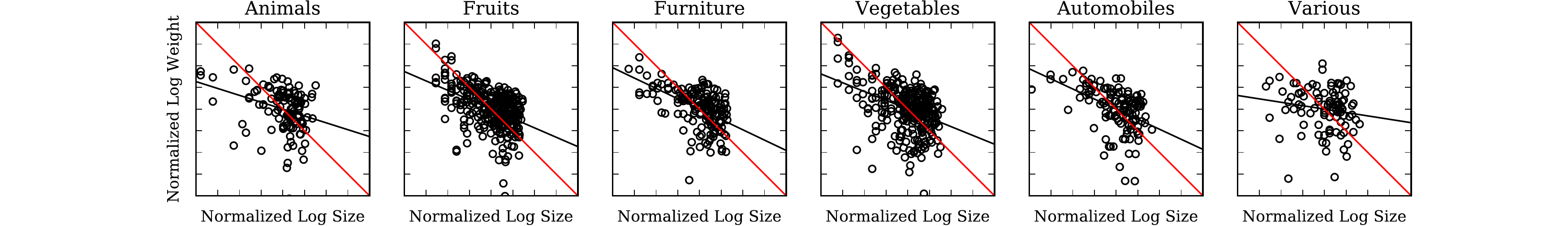}
\end{center}
\caption{Feature Size/Weight Relationships for Convolutional Neural Network Representations.} 
\label{perceptual-plots}
\vspace{-5px}
\end{figure*}

\section{Perceptual Hypothesis Spaces}

While evidence for the size principle seems apparent from studies of semantic hypothesis spaces, there has been no work attempting to verify the operation of the principle for concrete objects, especially with complex, real-world instances of these objects such as natural images. The featural representations of such instances are complex and include innumerable details not contained in semantic descriptions of the general case, rendering explicit feature descriptions difficult. Here, we offer a method to overcome this challenge by leveraging representations learned from deep neural networks.

\begin{figure}[!ht]
\begin{center}
\includegraphics[trim={0 0 0 0},clip,width=\linewidth,keepaspectratio]{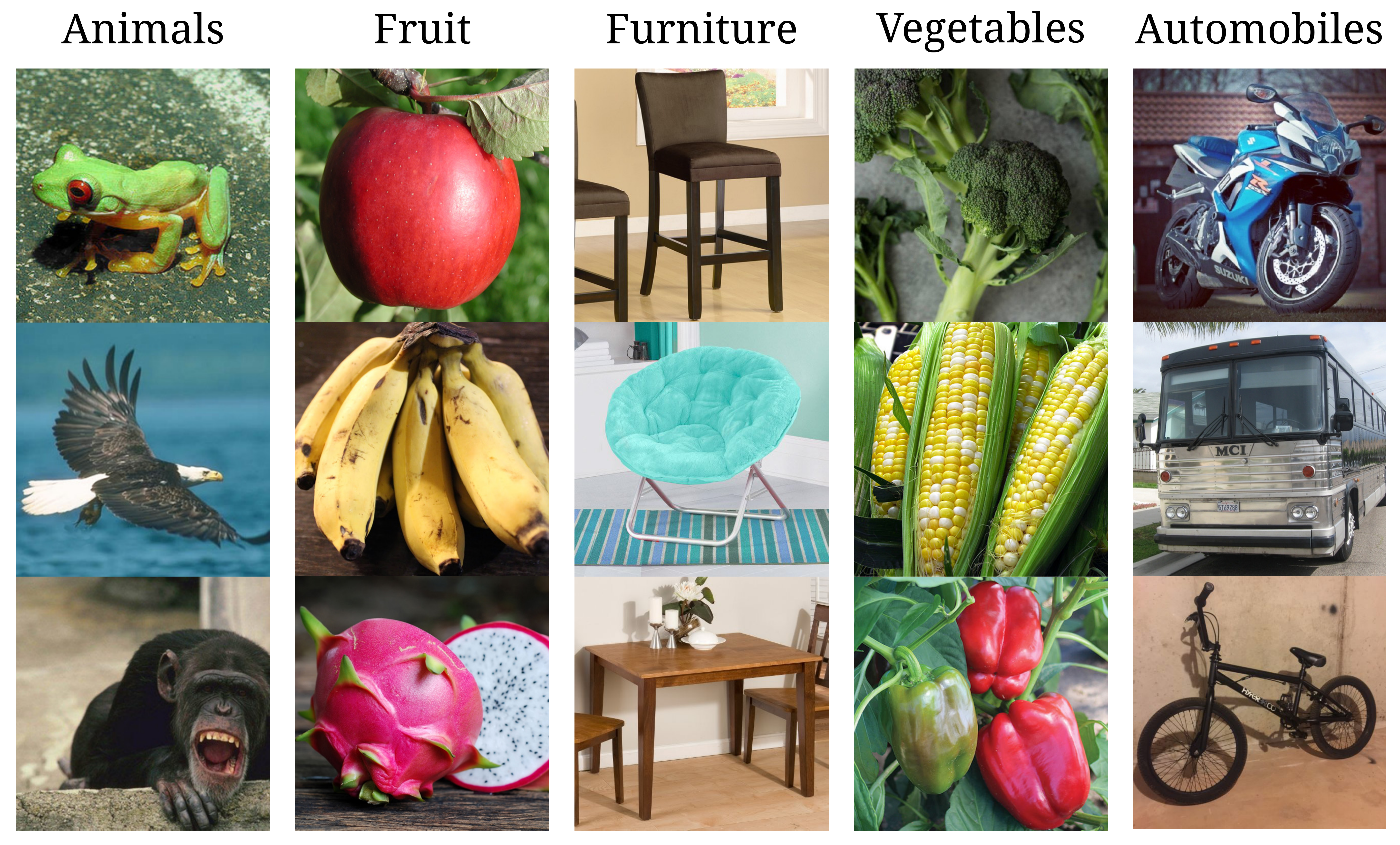}
\end{center}
\caption{Examples of stimuli from each of the 5 natural image categories} 
\label{image-examples}
\end{figure}

\subsection{Perceptual Features} Recent work \citep{peterson2016adapting} has provided evidence that deep image feature spaces can be used to approximate human similarity judgments for complex natural stimuli. For our analysis, we extracted image features from an augmented version of Alexnet with a binarized final hidden layer \citep{wu2015adjustable}. This allows for a comparison both to non-perceptual binary feature sets (i.e., features from the previous section) and non-binary perceptual feature sets (i.e., previous work on similarity prediction).

\subsection{Stimuli \& Data Collection} We obtained pairwise image similarity ratings for $5$ sets of $120$ images (animals, fruits, furniture, vegetables, vehicles) using Amazon Mechanical Turk, following \cite{peterson2016adapting}. Examples of images in each dataset are given in Figure \ref{image-examples}. The image sets represent basic level categories, with 20-40 subordinate categories in each.

Subjects rated at least $4$ unique pairs of images and we required that at least $10$ unique subjects rate each possible pair. Each experiment yielded a $120 \times 120$ similarity matrix.
\subsection{Analysis \& Results} As before, we computed the pairwise multiplication of each pair of rows in $\bf F$ ($120$ images $\times$ $4096$ neural features) and regressed this matrix onto the corresponding empirical similarity values. The resulting weights and feature sizes are plotted in Figure \ref{perceptual-plots} for each category, and the corresponding correlations are reported in Table \ref{CNN-corr-results}.

Like the previous semantic datasets, only a small portion of the total features obtained non-zero weights, although the average percentage was much smaller ($\sim 4$\%). Given that the full feature set is meant to characterize 1000 mostly qualitatively distinct categories from which they were learned \citep{deng2009imagenet}, whereas features from the semantic datasets were relevant only to the objects in each group, this discrepancy is to be expected. 

In all five datasets, correlation coefficients are moderate, negative, and significant at the $\alpha = 0.001$ level. Average Pearson and Spearman correlation was 0.42 in both cases. Variance explained in similarity matrices was comparable to previous work on predicting similarity from deep features, but was generally reduced given the constraint of binary features and non-negative weights.

\begin{table}
\begin{center}
\caption{Correlations between feature size and feature weight (Perceptual Dataset)}
\vspace{2mm}
\begin{tabular}{ l r r c r } 
\hline
\textbf{Set} & \textbf{Pearson} & \textbf{Spearman} & $\bm{FR}$ & $\bm{R^{2}_{MP}}$ \\ 
\hline
Animals & -0.32 & -0.34 & 122/4096 & 0.56 \\
Fruits & -0.43 & -0.44 & 302/4096 & 0.41 \\
Furniture & -0.48 & -0.51 & 170/4096 & 0.38 \\
Vegetables & -0.41 & -0.34 & 295/4096 & 0.45 \\
Automobiles & -0.46 & -0.49 & 125/4096 & 0.31 \\
\hline
\end{tabular}
\label{CNN-corr-results}
\end{center}
\end{table}

\section{Discussion}
We have attempted to provide a direct evaluation of the size principle in both semantic and perceptual hypothesis spaces. In some cases, the correlations we report using our method are weaker overall than those reported in past work \citep{navarro2010similarity}, but are consistently negative nonetheless. If anything, this discrepancy serves as a caution to trusting a single method for evaluating the size principle.

Across all datasets, variance explained in similarity judgments ranged from $.03$ to $.94$, however these fluctuations don't appear to vary systematically with the magnitude of the size principle effect, This may indicate that the size principle should emerge with respect to both ``good" and ``bad" feature sets, so long as they are related to the objects and vary in their inclusiveness.

Furthermore, it appears that the size principle can be shown to operate in more ecologically valid stimulus comparisons such as visual image pairs. In cases such as these, the specific visual details of the image are relevant, and our feature sets derived from convolutional neural networks included only these features. There may be hundreds of small visual details that are only present in novel instantiations of familiar objects that we encounter on a daily basis and that actually represent the more abstract concepts used in semantic datasets. These results may also have implications for the method of estimating human psychological representations recently proposed by \cite{peterson2016adapting}. In this work, it was shown that human similarity judgments for natural images can be estimated by a linear transformation of deep network features, and the current results imply that this transformation is perhaps partly accounted for by the size principle. This finding may lead to better methods for approximating complex human representations based on psychological theories. 

It is apparent from the $FR$ columns of each table that few of the total features were used in the actual models. This may be due in part to useless features, or features associated with too many or too few objects. It may also be due to multicollinearity in our feature matrices (some columns are linear combinations of others). These are unique consequences of using a regression model. For this reason, our method may be less susceptible to over-representing certain features that are redundant. On the other hand, the size principle is meant to address the problem of redundant hypotheses directly, and it may be an undesirable property of our model that these hypotheses are eliminated through other means, which is perhaps the cost of direct estimation of the weights in the additive clustering framework. In any case, this variability in the amount of non-redundant features does not appear to co-vary with the size principle in any systematic way.

The only notable discrepancy between our results and the predictions of the size principle is the variation in the magnitude of the negative slopes obtained, which does not appear to depend on model performance, number of features, or even aspects of the dataset groups or individual datasets. Semantic dataset group 2 had more large slopes (e.g., SIMLEX) than group 1, but also had many small slopes. Similar datasets from group 1 (e.g., Fruit and Vegetables) had fairly dissimilar slopes, and nearly identical datasets from group 2 (e.g., Fruit1 and Fruit2) had widely varying slopes. Prototypicality doesn't seem to matter either, since Fruit1 and Vegetables1 have smaller slopes than Fruit2 and Vegetables2, but Clothing1 and Vehicles1 have larger slopes than Clothing 2 and Vehicles2. Furthermore, we can find examples of both natural and artificial stimuli with comparable slopes. For these reasons, it is unclear what the source of these deviations could be. It is possible that certain experimental contexts encourage a focus on certain featural comparisons that can be represented by a weighting of our feature sets, and so still allow for good model fit. Alternatively, it may be an artifact of the weight estimation algorithm, in which case it will be useful to compare alternative methods.

Our results provide broad evidence for the size principle regardless of the assumptions that are employed to derive it. Thus, the size principle is a good candidate for a second universal law of generalization, and can be motivated both by rational theories based on strong sampling and feature learning. Further, a $\frac{1}{|h|}$ law can provide a solid basis for generalizing from multiple instances, a behavior that we should expect to find in any intelligent agent, anywhere in the universe.


\bibliographystyle{apacite}

\setlength{\bibleftmargin}{.125in} 
\setlength{\bibindent}{-\bibleftmargin}
\vspace{-1mm}
\bibliography{CogSci_Template}

\end{document}